\pdfoutput=1

\documentclass[11pt]{article}

\usepackage[final]{coling}

\usepackage{times}
\usepackage{latexsym}
\usepackage{multirow}
\usepackage{booktabs}
\usepackage[T1]{fontenc}
\usepackage{tipa}

\usepackage[utf8]{inputenc}
\usepackage{amsmath}
\usepackage{arydshln}
\usepackage{amssymb}

\usepackage{adjustbox}

\usepackage{pifont}
\newcommand{\cmark}{\ding{51}}%
\newcommand{\xmark}{\ding{55}}%

\usepackage{microtype}

\usepackage{inconsolata}

\usepackage{graphicx}

\title{Cross-Dialect Information Retrieval: Information Access\\in Low-Resource and High-Variance Languages}

\author{Robert Litschko$^{\mathbf{1,2}}$ ~~ Oliver Kraus$^{\mathbf{1}}$ ~~ Verena Blaschke$^{\mathbf{1,2}}$ ~ Barbara Plank$^{\mathbf{1,2}}$\\
$^{\mathbf{1}}$ MaiNLP, Center for Information and Language Processing, LMU Munich, Germany \\
$^{\mathbf{2}}$ Munich Center for Machine Learning (MCML), Munich, Germany\\
\texttt{\{robert.litschko, o.kraus2, verena.blaschke, b.plank\}@lmu.de}}

\begin{document}
\maketitle

\begin{abstract}
A large amount of local and culture-specific knowledge (e.g., people, traditions, food) can only be found in documents written in dialects. While there has been extensive research conducted on cross-lingual information retrieval (CLIR), the field of cross-dialect retrieval (CDIR) has received limited attention. Dialect retrieval poses unique challenges due to the limited availability of resources to train retrieval models and the high variability in non-standardized languages. We study these challenges on the example of German dialects and introduce the first German dialect retrieval dataset, dubbed WikiDIR, which consists of seven German dialects extracted from Wikipedia. Using WikiDIR, we demonstrate the weakness of lexical methods in dealing with high lexical variation in dialects. We further show that the commonly used zero-shot cross-lingual transfer approach with multilingual encoders does not transfer well to extremely low-resource setups, motivating the need for resource-lean and dialect-specific retrieval models. We finally demonstrate that (document) translation is an effective way to reduce the dialect gap in CDIR. 
\end{abstract}

\section{Introduction}

Cross-lingual information retrieval (CLIR) is the task of retrieving documents written in a language different from the query language. The promise of CLIR is to make information accessible across language boundaries. One of its main challenges lies in bridging the \textit{lexical gap between languages} \citep{berger2000bridging}, which is caused by the fact that different languages use different vocabularies. This problem has been extensively studied in the context of ad-hoc news retrieval \citep{braschler2003clef,hc4} and Wikipedia passage retrieval \citep{sasaki-etal-2018-cross,sun-duh-2020-clirmatrix,ogundepo-etal-2022-africlirmatrix,li-etal-2022-museclir}, among others.  

\begin{figure}[t!]
    \centering
    \includegraphics[width=0.8\linewidth]{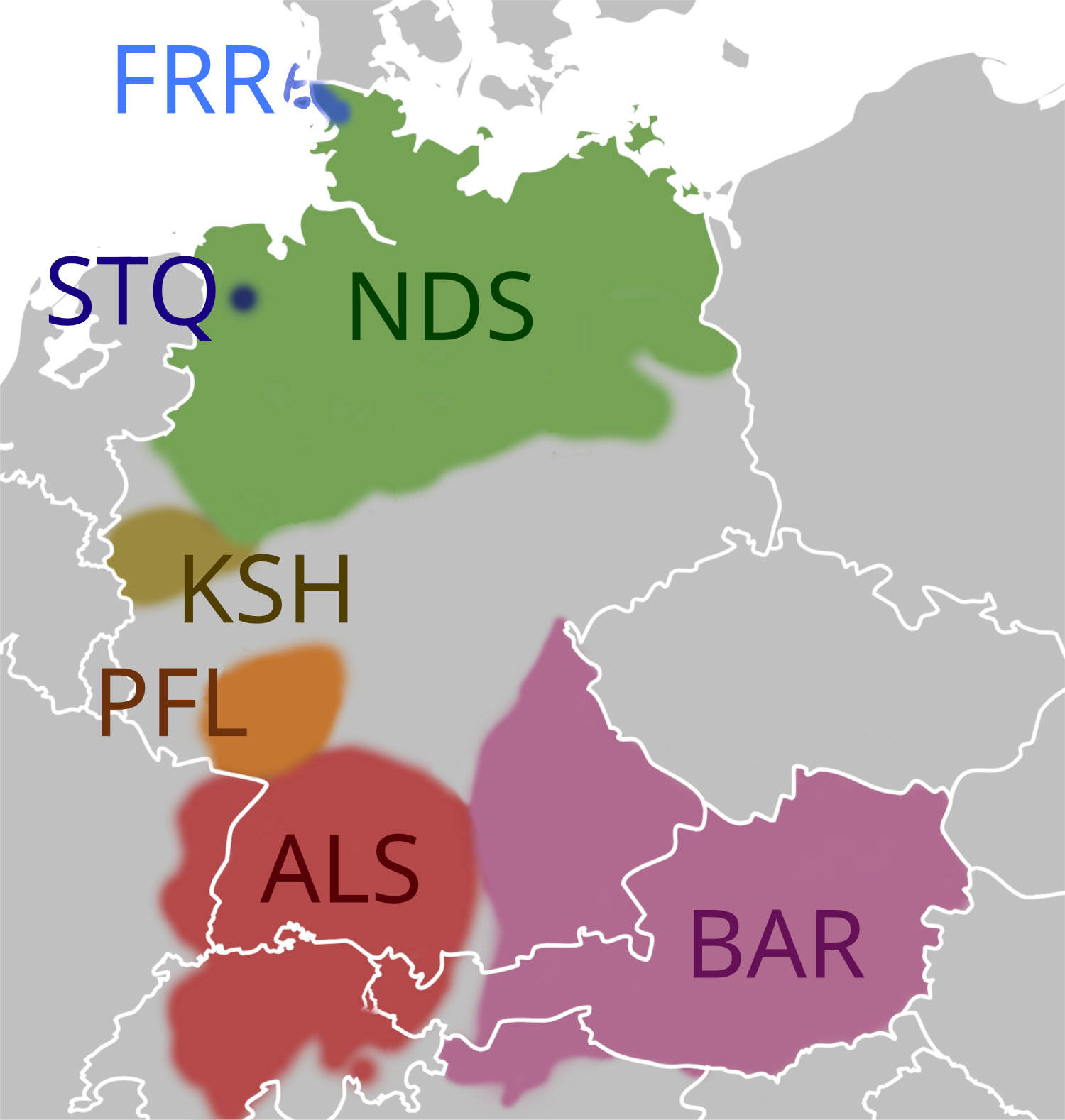}
    \caption{Approximate locations of German regional languages and dialects studied in this work: Low German (nds), North Frisian (frr), Saterfrisian (stq), Ripuarian (ksh), Rhine Franconian (pfl),  Alemannic (als), and Bavarian (bar). Image based on a map of Europe by Marian Sigler, \href{https://creativecommons.org/licenses/by-sa/3.0/}{CC BY-SA 3.0}.}
    \label{fig:map}
\end{figure}

Relevant information may, however, not only be written in a different language but also in a regional dialect or language variant. This is particularly true for culturally-related concepts such as local traditions, individuals, and locations. In this work, we focus on cross-dialect information retrieval (CDIR) with German regional languages and dialects (see Figure~\ref{fig:map}). The \textit{lexical gap between dialects} in CDIR is caused by orthographic variations and different regional expressions, resulting from the lack of standardization. For example, the German term \textit{München} (``Munich'') is also spelled as \textit{Münche} (als), \textit{Minga} (bar), \textit{Münsche} (ksh) and \textit{Minche} (pfl). In addition to variations found in different dialects, there are also variations that arise from subdialects. 
The dialect variation renders lexical retrieval methods such as BM25 \citep{robertson2009probabilistic} and CLIR approaches based on machine translation \citep{lignos-etal-2019-challenges,artetxe-etal-2023-revisiting} less effective, since they only match a single variation. 
At the same time, many dialects only have a small digital footprint. All dialects used in this work are categorized as low-resource languages \citep{joshi-etal-2020-state}. It is unclear how well  resource-lean methods such as zero-shot cross-lingual transfer paradigm based on pre-trained language models \citep{macavaney2020teaching,shi-etal-2020-cross,litschko2022cross,litschko-etal-2023-boosting} or large language models \citep{sun-etal-2023-chatgpt} perform on CDIR.  
While dialect variation has recently gained traction in the NLP community \citep{faisal-etal-2024-dialectbench}, it is still understudied in the context of retrieval. The only recent work we are aware of is by \citet{chari2023effects}, who compare retrieval performance between British English and American English spelling variants found in MS MARCO \citep{bajaj2016ms}, and \citet{valentini2024messirve}, who compare retrieval models on different Spanish variants. 
To close this gap, we introduce WikiDIR, a cross-dialect retrieval dataset based on German dialect Wikipedias. 
WikiDIR includes queries in standard German and documents in seven regional language variants, which we will refer to as dialects throughout the rest of this paper. 
Different from prior work, we specifically focus on orthographic and lexical variation on languages that are non-standardized and have limited resources. Common approaches to derive queries from Wikipedia include using article titles \citep{sun-duh-2020-clirmatrix,ogundepo-etal-2022-africlirmatrix} or the first sentence \citep{sasaki-etal-2018-cross}. These approaches, however, fall short of capturing full spectrum of dialect variation. To address this limitation, and as another contribution, we manually annotated entity mentions in for five dialects and built dictionaries that contain different spelling variants checked by native speakers. 

In summary, we are the first to study dialect-specific lexical variation in a low-resource and high-variation setting. We release our WikiDIR dataset,  dialect dictionaries, as well as our annotations and source code.\footnote{\href{https://github.com/mainlp/WikiDIR}{https://github.com/MaiNLP/WikiDIR}}
We address the following research questions: \textbf{RQ1:} How well do lexical and neural retrieval models perform on CDIR? \textbf{RQ2:} To what extent can we improve their performance by specializing models for dialects models by means of continual pretraining? \textbf{RQ3:} How does the lack of standardization in dialects, i.e., the dialect gap, affect retrieval performance? \textbf{RQ4:} How well do large language models fare when employed as dialect translation models?

\section{Related Work}

\paragraph{Wikipedia-based test collections.} Wikipedia has been extensively used as a resource for training and evaluating cross-lingual retrieval models in the context of ad-hoc retrieval \citep{schamoni-etal-2014-learning,sasaki-etal-2018-cross,sun-duh-2020-clirmatrix,li-etal-2022-museclir,ogundepo-etal-2022-africlirmatrix} and cross-lingual question answer retrieval \citep{roy-etal-2020-lareqa,asai-etal-2021-xor}. Queries and documents are typically extracted from article titles and article texts. Most CLIR datasets derived from Wikipedia obtain relevance labels by propagating synthetic monolingual relevance labels through inter-language links. They can be broadly grouped by how monolingual labels are derived. One approach is to synthesize relevance labels based on (mutual) links between Wikipedia articles \citep{schamoni-etal-2014-learning,sasaki-etal-2018-cross,frej-etal-2020-wikir,li-etal-2022-museclir}. The second approach, used in more recent benchmarks such as CLIRMatrix \citep{sun-duh-2020-clirmatrix} and AfriCLIRMatrix \citep{ogundepo-etal-2022-africlirmatrix}, performs monolingual lexical retrieval and obtains relevance labels by discretizing BM25 retrieval scores. In this work, we adopt and extend this approach (cf. Section~\ref{s:data}). The use of synthetic queries and relevance labels has been criticized in \citep{zhang-etal-2023-miracl}, because they bias CLIR evaluation towards a lexical notion of relevance, rather than semantic relevance. Other Wikipedia-based CLIR datasets such as Mr.TyDi \citep{zhang-etal-2021-mr} and XOR-QA \citep{asai-etal-2021-xor} are centered around QA and require a more sophisticated query interpretation. Our focus, however, lies in \textit{dialect-specific lexical variation} that can be attributed to the lack of standardization. 

\begin{table*}[ht!]
\centering
\small 
\begin{tabular}{l r r r r r r r r r}
 \toprule
 & \multicolumn{4}{c}{Wikipedia articles (total)} & &  \multicolumn{4}{c}{Relevance assessments (avg)} \\ \cmidrule(lr){2-5} \cmidrule(lr){7-10} 
 &  \#Train & \#Dev & \#Test & \#Analysis & Total & \#Train & \#Dev & \#Test & \#Analysis \\ \midrule
German & 64,253 & 8,032 & 8,031 & -- & 80,316 & 4.8 & 5.0 & 2.4 & -- \\ \midrule 
Low German (nds) & 58,901 & 7,382 & 7,367  & 470 & 74,120 & 12.8 & 12.2 & 6.5 & 269.9 / 315.9  \\
Alemannic (als) &  17,092 & 2,121 & 2,146 & 4,639 & 25,998 & 13.4 & 13.2 & 6.9 & 35.0 / 60.0 \\
Bavarian (bar) & 17,974 & 2,234 & 2,244 & 718 & 13,170 & 8.9 & 8.8 & 4.4 & 77.1 / 91.6 \\
North Frisian (frr) & 12,550 & 1,594 & 1,565  & -- & 15,709 & 9.9 & 10.6 & 4.7 & --  \\
Saterfrisian (stq) & 2,710 & 342 & 339  & -- & 3,391 & 9.5 & 8.9 & 4.5 & --  \\
Ripuarian (ksh) & 1,800 & 228 & 222  & 210 & 2,460 & 6.6 & 6.5 & 3.4 & 19.9 / 30.3 \\
Rhine Franconian (pfl) & 1,971 & 244  & 250 & 157 & 2,622 & 3.6 & 3.2 & 1.8 & 23.0 / 36.8 \\ \midrule
\textbf{Total/Average} & 177,251 & 22,177 & 22,164 & 6,194 & 227,786 & 9.2 & 9.1 & 4.6 & 85.0 / 106.3 \\ 
\bottomrule
\end{tabular}

\caption{WikiDIR statistics. We report the number of Wikipedia articles, which corresponds to the number of queries and documents. We also show for each dialect the average number of relevance annotations. In addition to the standard train-dev-test splits, we also use a held-out analysis split consisting of queries for which we found at least one dialect-specific lexical variation (see Section \ref{sec:dialectvariatoin}). To quantify the lexical dialect gap, we provide relevance assessments without (\xmark) / with (\cmark) documents containing dialect variations. 
}
\label{tab:stats}
\end{table*}

\paragraph{Cross-lingual transfer.} Two common ways to bridge the language gap in CLIR include the so-called translate-train and translate-test approach \citep{saleh-pecina-2020-document,bonifacio2021mmarco,artetxe-etal-2023-revisiting}. The drawback of machine-translated test collections is that they not only suffer from translationese \citep{zhang-toral-2019-effect,zhao-etal-2020-limitations} but are also culturally biased towards the language from which they are translated from \citep{hecht2010tower,gutierrez-etal-2016-detecting}. Due to their low-resource nature, dialects studied in this work are not supported by neural machine translation such as Google Translate or NLLB \citep{costa2022no}. Unlike German dialects, standard German is regarded as a high-resource language \citep{joshi-etal-2020-state}. We therefore study the commonly-adopted zero-shot cross-lingual transfer paradigm \citep{hu2020xtreme,liang-etal-2020-xglue} for CDIR. Specifically, we study the zero-shot transfer based on multilingual BERT \citep{devlin-etal-2019-bert} and the recently proposed reranking approach based on large language models (LLMs) \citep{sun-etal-2023-chatgpt}.

\paragraph{Dialect retrieval.} Most of the dialects studied in this work are often overlooked in prior work. For example, \citet{artemova-plank-2023-low} study bilingual lexicon induction (i.e., word-level CDIR) on Bavarian and Alemannic. On sentence-level CDIR, WikiMatrix \citep{schwenk-etal-2021-wikimatrix} covers Low German and Bavarian. \citet{vamvas-etal-2024-modular} investigate continual pretraining for Swiss German sentence retrieval. On passage-level retrieval, \citet{chari2023effects} separate American from British spelling variants found in MS MARCO and find that neural retrieval models, despite showing a clear preference towards the former, exhibit robust retrieval performance. Most recently, \citet{valentini2024messirve} present MessIRve, a retrieval benchmark with queries sourced from different Spanish-speaking regions using Google's autocomplete API and documents corresponding to Spanish Wikipedia articles. Finally, the CLIRMatrix benchmark \citep{sun-duh-2020-clirmatrix} uses English queries and covers three out of four dialects studied in this work (als, bar, nds). Different from prior work, we investigate the CDIR performance in truly low-resource settings and contribute bilingual dictionaries containing diverse lexical and orthographic dialect variants, which allow us to quantify the impact of the dialect gap.

\section{WikiDIR Benchmark}
\label{s:data}

Our benchmark includes data from seven regional languages and dialects spoken in Germany: 
North Frisian (wiki code: \href{https://frr.wikipedia.org}{frr}), 
Saterfrisian (\href{https://stq.wikipedia.org}{stq}), 
Low German (\href{https://nds.wikipedia.org}{nds}), 
Ripuarian (\href{https://ksh.wikipedia.org}{ksh}), 
Rhine Franconian (\href{https://pfl.wikipedia.org}{pfl}), 
Alemannic (\href{https://als.wikipedia.org}{als}), 
and Bavarian (\href{https://bar.wikipedia.org}{bar}).
Most of these wikis contain multiple dialect subgroups.

\subsection{Dataset Pipeline}
\label{sec:pipeline}
\begin{figure*}[ht]
    \centering
    \includegraphics[width=0.95\linewidth]{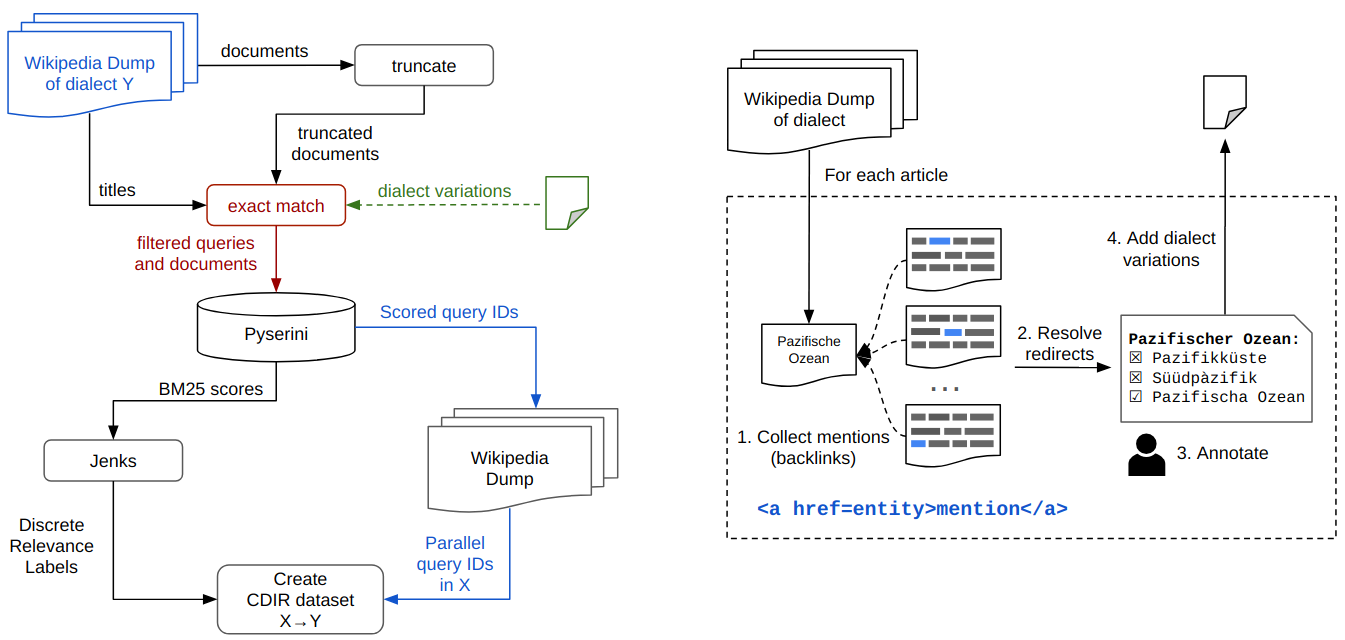}
    \caption{\textbf{Left}: Dataset pipeline used to create WikiDIR. 
    Our approach deviates from CLIRMatrix \citep{sun-duh-2020-clirmatrix} in three aspects (highlighted in color). First, we only label documents as relevant if they contain an exact match of the query. Second, we use dialect dictionaries to also label documents as relevant if they contain any dialect variation of the query. Third, we use inter-language links to replace dialect queries by their German translation, rather than propagating labels on the document side. \textbf{Right}: Annotation pipeline used to build dialect dictionaries containing orthographic and lexical variations. We extract candidates of dialect variations (entity mentions) from incoming links. Correct variations are human-annotated and added to the dictionary. 
    We provide for a set of held-out queries (analysis split) relevance assessments with (\cmark) and without (\xmark) documents containing dialect variations. 
    }
    \label{fig:annotation-pipeline}
\end{figure*}

We now describe our dataset pipeline used to create WikiDIR, which closely resembles the approach proposed in CLIRMatrix \citep{sun-duh-2020-clirmatrix}. Figure~\ref{fig:annotation-pipeline} shows our dataset pipeline and Table~\ref{tab:stats} summarizes key statistics for each dialect.

\paragraph{Queries and documents.} We use Wikipedia article titles as queries and treat as documents the first 200 extracted tokens from the article texts. We specifically use the display title of each dialect article, which is different from the canonical title written in standard. Since our focus is on lexical variation between German and its dialects, we exclude all articles where the queries consist solely of numbers. Many Wikipedia articles contain lexical shortcuts, i.e., references to the official German spelling in the first sentence: 
\textit{Minga (amtli: \underline{München}) [\textprimstress{}m\i{}\textipa{N}(:)\textturna] is d'Haptstod vo Bayern.}.\footnote{``Munich (officially \textit{München}) is the capital of Bavaria.''}
To address this issue and the risk for the task of becoming a mere (monolingual) lexical keyword matching task, we use German article titles to eliminate such lexical shortcuts. The example thus becomes: 
\textit{Minga [\textprimstress{}m\i{}\textipa{N}(:)\textturna] is d'Haptstod vo Bayern.}

\paragraph{Relevance labels.} 
Obtaining human-labeled relevance annotations is expensive and does not scale to a large number of languages. Because of this, we follow prior work \citep{sun-duh-2020-clirmatrix,ogundepo-etal-2022-africlirmatrix} and opt for synthetic relevance labels obtained from lexical retrieval scores. That is, we first obtain monolingual relevance labels from discretized lexical retrieval scores. For this, we use the BM25 implementation provided by the Pyserini framework \citep{lin2021pyserini}. This approach, however, falls short in dealing with multi-term queries containing 
terms that are identical in the dialect and German. Consider for example the Alemannic query \textit{Kanton Lozärn} (``Canton of Lucerne''). Under the bag-of-words assumption a BM25 retriever would first return many documents containing the German term \textit{Kanton}. Labeling documents with partial lexical matches as relevant would therefore lead to many false positives, which skews the retrieval evaluation towards standard German rather than dialect retrieval. Because of this, we only score documents with BM25 (and thus label as relevant) if they contain an exact phrasal match. To obtain monolingual labels, we follow the approach proposed by \citet{sun-duh-2020-clirmatrix} and use min-max normalization to normalize all BM25 values to the unit range and discretize them into the labels 1--5 using Jenks natural breaks optimization \citep{mcmaster2002history}. Documents extracted from the same articles as queries are labeled with 6. Finally, to obtain cross-lingual relevance labels, we use inter-language links and replace dialect titles (i.e., dialect queries) with German titles.

\subsection{Dialect Variation}
\label{sec:dialectvariatoin}
Relying on Wikipedia titles to annotate the relevance of documents does not account for the dialect variance resulting from the lack of standardization. In this work, we investigate dialect variation from two perspectives: (1) orthographic and lexical variation and (2) variation within subdialects. 

\paragraph{Orthographic and lexical variation.} 
Broadly speaking, dialect variation can be categorized into orthographic differences based on different regional pronunciations (e.g., \textit{Minga} and \textit{Mincha} referring to Munich) and different regional word choices (e.g., \textit{Brötchen}, \textit{Weckli}, and \textit{Semml} referring to bread rolls) \citep{barbour1998variation}. 
While lexical and morphological variation in German dialects and regional languages have been well-documented, e.g., by \citet{ada}, there exist very few publicly available resources. Recent efforts on lexical resources include, e.g., building crowdsourcing translations for Bavarian \citep{burghardt-etal-2016-creating} and lexicon induction via rule extraction \citep{millour-fort-2019-unsupervised}. To study the impact of variation on CDIR, we build entity-centric dialect dictionaries from human annotations on five dialects. The annotation pipeline is illustrated in Figure~\ref{fig:annotation-pipeline}. For each entity, we first extract all backlinks from the raw Wikipedia dumps. The anchor link text typically corresponds to entity mentions written in dialect, while the href attribute stores the standard German spelling. We use the Wikimedia REST API to resolve redirects and obtain normalized dialect article titles. This way, we collect for each dialect title (i.e., query) all mentions. As shown in Table~\ref{tab:annotations}, we extracted over 53k mentions for 9.9K entities in five dialects. 

\def\arraystretch{1.2}
\begin{table}
\centering
\small 
\begin{tabular}{l r r r r} \toprule
& \multicolumn{2}{c}{Anchor links} & \multicolumn{2}{c}{Dictionary} \\ \cmidrule(lr){2-3} \cmidrule(lr){4-5}
Dialect & Entities & Mentions & Entities & Variants \\ \midrule
nds & 1,653 & 6,579 & 484 & 1.55 \\
als & 6,440 & 38,671 & 4,684 & 3.11 \\
bar & 1,402 & 6,289 & 721 & 2.12 \\
ksh & 232 & 1,056  & 211 & 2.92 \\
pfl & 182 & 862  & 157 & 2.79 \\ \midrule
\textbf{Total/Avg} & 9,909 & 53,457 & 6,257 & 2.50 \\
\bottomrule
\end{tabular}
\caption{Number of entities and entity mentions (i.e., candidates) per dialect, as well as the average number of variants per entity after annotation.}
\label{tab:annotations}
\end{table}

Many entity mentions do not correspond to lexical variations in a strict sense. Consider for example the Alemannic dialect title \textit{Schtadt} (``city''), while \textit{Schtatt} and \textit{Sdadd} refer to correct orthographic variations, the term \textit{Schtädt} refers to its plural form (``cities''). Next to grammatical number, we also consider deviations in grammatical gender, grammatical case and word classes. Lexical normalization is typically done through stemming or lemmatization. However, such lexical analyzers do not yet exist for German dialects. Also, entity mentions often correspond to abbreviations or semantically related terms. This is for example the case for the term \textit{Zitig} (``newspaper''), where \textit{Zittung} is a correct variation while the compound noun \textit{Daageszitig} refers to a daily newspaper. Since we do not want to inflate the dialect gap with different preprocessing steps, we only focused on dialect variations in a strict sense (regional word choices and spelling variations), rather than variations that also occur in standard German (different word inflections, synonyms, paraphrases). These variations were annotated by native German speakers with a computational linguistics background. We calculated the inter-annotator agreement on a sample of three dialects and 50 queries in terms of Cohen's Kappa \citep{cohen1960coefficient}. The Kappa values for \textit{pfl} ($\kappa=0.83$), \textit{bar} ($\kappa=0.87$), and for \textit{als} ($\kappa=0.80$) all show high agreements. Notably, the resulting dialect dictionaries vary considerably in size (cf. Table~\ref{tab:annotations}). This is to some extent due to the fact that smaller Wikipedias have fewer editors and fewer links between articles \cite{arora2024orphan}. On average, our dictionaries store for each dialect query 2.5 dialect-specific variations. 

We use our dictionaries to create a held-out analysis split consisting of the subset of queries for which we found at least one dialect variation (Figure \ref{fig:annotation-pipeline}, right). To measure the impact of dialect variations on retrieval (see Tables~\ref{tab:analysis_results} and \ref{tab:translated-results}), we create two relevance assessments for our analysis split. In the first version (\xmark) we label documents containing the query (i.e., dialect title) as relevant (e.g., Bavarian documents containing the term \textit{Minga}). In the second version (\cmark), we additionally label all documents that contain any dialect variation as relevant (\textit{Münch'n}, \textit{Minkcha}, \textit{Minkn}, \textit{Minchn}, \textit{Mingna}, \textit{Minkhn}, \textit{Münchn}). From here, we obtain cross-lingual relevance labels as described in Section \ref{sec:pipeline}. We make all dictionaries, together with WikiDIR, publicly available for future uptake.

\paragraph{Dialect groups.} 
To investigate language variation within individual wikis, we use the subset of articles tagged with fine-grained dialect information.
We use the dialect groupings by \citet{wiesinger1983einteilung} and \citet{walker2001nordfriesisch} to assign articles (i.e., subdialects) to comparable dialect subgroups.
This is possible for Bavarian (North, Central, and South Bavarian), Alemannic (Swabian, and Low, Central, High, and Highest Alemannic), and North Frisian (Insular and Mainland North Frisian).
We do not divide Ripuarian or Rhine Franconian into subgroups, since they each only contain one of \citeposs{wiesinger1983einteilung} groups.
There are no dialect tags available for Saterfrisian, and too few for meaningful analysis in the case of Low German.

\section{Models}
We benchmark retrieval models that are  widely-used in monolingual and cross-lingual information retrieval. Using WikiDIR, we investigate how well they perform on CDIR and to what extent they are able to deal with dialect-specific variation.

\subsection{Multi-Stage Retrieval} 
We first investigate the multi-stage retrieval paradigm, which has been proposed by \citet{nogueira2019multi}. We use BM25 as a first-stage retriever \citep{robertson2009probabilistic} implemented in the Pyserini framework \citep{lin2021pyserini}. We follow prior work and re-rank the top 100 documents using MonoBERT \cite{macavaney2019cedr,litschko2022cross}. 
MonoBERT is a cross-encoder model that treats relevance prediction as a sequence-pair classification task \citep{nogueira2019passage}. Specifically, MonoBERT concatenates query document pairs \texttt{[CLS] Query [SEP] Doc [SEP]} and predicts the relevance score of each document independently of other documents. Here, we opt for multilingual BERT (mBERT, \citealp{devlin-etal-2019-bert}) as the encoder model. We use the pairwise hinge loss  
\citep{macavaney2019cedr,sun-duh-2020-clirmatrix} and compare MonoBERT in two settings. We first fine-tune it on the respective training splits of WikiDIR. Given the low-resource nature of many dialects, we are also interested in the zero-shot transfer from German. To this end, we also train a MonoBERT model only on the standard German portion of WikiDIR.

\subsection{Re-Ranking with LLMs}
\citet{sun-etal-2023-chatgpt} demonstrate the effectiveness of commercial LLMs on zero-shot re-ranking tasks. To this end, an LLM is prompted with the query and the respective retrieved documents. The task of the model is then to generate a new ranking of the documents based on their relatedness to the query. As a large number of retrieved documents can quickly fill up the context window of LLMs, the authors propose a sliding-window approach for re-ranking. This method iteratively re-ranks subsets of documents, moving backwards through the list until all documents have been evaluated.
Building on this work, \citet{zhang2023rank} demonstrate the effectiveness of this approach on open-source LLMs. 
In our work, we therefore adapt this approach to evaluate its effectiveness on low-resource dialect data. For our experiments we use the Meta-Llama-3-8B-Instruct\footnote{\url{https://huggingface.co/meta-llama/Meta-Llama-3.1-8B-Instruct}\label{footnote-llama3}} model as a state-of-the-art open-source LLM.
We use the same prompt template for re-ranking as presented by \citet{sun-etal-2023-chatgpt}.  
\begin{table}[!htbp]
\small
    \centering
    \colorbox{gray!8}{
    \begin{tabular}{@{}p{7.3cm}}
    ============= \textsc{LLM-Reranking} =============\\\\

    \textbf{system:}
You are RankGPT, an intelligent assistant that can rank passages based on their relevancy to the
query. \\\\

\textbf{user}:
I will provide you with {{num}} passages, each indicated by number identifier []. Rank them based on their relevance to query: \{\{query\}\}. \\\\

\textbf{assistant:}
Okay, please provide the passages. \\\\

\textbf{user:}
{[1]} \{\{passage\_1\}\} \\\\

\textbf{assistant:}
Received passage [1] \\\\

\textbf{user:}
{[2]} \{\{passage\_2\}\}  \\\\

\textbf{assistant:} Received passage [2] \\\\

(more passages) ...

\\\\
\textbf{user:} \\
Search Query: \{\{query\}\}. \\
Rank the {{num}} passages above based on their relevance to the search query. The passages should be listed in descending order using identifiers, and the most relevant passages should be listed first, and the output format should be [] > [], e.g., [1] > [2]. Only response the ranking results,
do not say any word or explain.
    
    \end{tabular}}
\label{tab1}
\end{table}
We use greedy decoding with a sliding window size of 16 and a step size of 8. This is a scaled-down version of the settings used by \citet{sun-etal-2023-chatgpt}, to ensure that we do not exceed the context window of our model.
Similar to MonoBERT, we re-rank the top 100 documents retrieved by BM25. If BM25 returns less than 100 documents, they are not filled up with random documents.

\subsection{ColBERTv2}
Naturally, the performance of multi-stage retrieval models is bound by the recall of the first-stage ranker. \citet{khattab2020colbert} introduce the ColBERT model, which achieves effective and efficient retrieval with a late interaction mechanism between query and document. Documents and queries are separately encoded, which are then compared using a MaxSim operator, allowing for token-level matching during retrieval. ColBERT can be used for both re-ranking and full retrieval.
\citet{santhanam-etal-2022-colbertv2} introduce ColBERTv2, which decreases the stored index size while providing higher quality retrieval results.

We train ColBERTv2 models using mBERT\footnote{\url{https://huggingface.co/google-bert/bert-base-multilingual-uncased}} models. All models are used for full retrieval on the datasets, rather than re-ranking. Due to our comparatively small datasets, we modify the standard hyperparameters by \citet{santhanam-etal-2022-colbertv2} by reducing the batch size to 16 and not using warm-up steps. Additionally, as with MonoBERT, we train a ColBERT model only on the standard German portion of WikiDIR to investigate its zero-shot cross-lingual transfer performance.

\subsection{Continual Pretraining of MLMs}
\label{sec:continual-pretraining}

\citet{senel-etal-2024-kardes} show that by jointly pretraining a masked language model on a low resource language together with a related high resource language, they can improve cross-lingual transfer. Specifically, they introduce Bilingual Alternating LM-ing (BALM) and Bilingual Joint LM-ing (BJLM). 
For BALM, the model is trained on monolingual batches from a high-resource and a low-resource language in an alternating fashion. 
For BJLM, each batch contains an equal ratio of high-resource and low-resource instances. In this work, we pretrain multiple masked language models using BALM and BJLM to investigate whether continual pretraining can improve the retrieval performance for low-resource dialects. 

Given that our dataset consists of multiple closely related dialects, we also experiment with Multilingual Alternating LM-ing (MALM) and Multilingual Joint LM-ing (MJLM). For MALM, the model is trained on a batch of German instances, followed by a batch from each of the seven dialects in our dataset, until all instances from the largest dataset have been processed. For MJLM, each batch contains an equal ratio of high-resource samples (i.e., German) and samples from all seven dialects in our dataset. This approach ensures that the model is exposed to a balanced ratio of the high-resource language and all low-resource dialects simultaneously during training.

We continually mBERT using BJLM, BALM, BALM and MALM with the train splits of all seven dialects of WikiDIR. We train all models with a learning rate of 1e-4 and a batch size of 128 for one epoch. All processed documents are split into chunks of 128 tokens.

\subsection{Document Translation with LLMs}
\label{sec:document-translation}
\citet{adeyemi-etal-2024-zero} investigate the effectiveness of using LLMs as translation models in the context of CLIR with African languages. In this work, we investigate the suitability of the Meta-Llama-3-8B-Instruct model as a dialect translation model. We use greedy decoding to translate documents from different dialects to standard German. The translated documents are used (1) as training data for ColBERTv2 and MonoBERT models, and (2)~at retrieval time. We follow \citet{adeyemi-etal-2024-zero} and use the following prompt template:

\begin{table}[!htbp]
\small
    \centering
    \colorbox{gray!8}{
    \begin{tabular}{@{}p{7.3cm}}
    ========= \textsc{Translate Train Prompt} =========\\\\

    Document: \{document\} \\\\

    Translate this document to German. Only return the translation, do not say any other word. \\
    \end{tabular}}
\end{table}

\def\arraystretch{1.2}
\setlength{\tabcolsep}{10pt}
\begin{table*}[ht!]
\centering
\small 

\begin{tabular}{l c c c c c c c c}
\toprule
Model & \textit{als} & \textit{bar} & \textit{nds} & \textit{frr} & \textit{stq} & \textit{ksh} & \textit{pfl} & \textbf{avg} \\ \midrule 
\texttt{BM25} & 0.750 & 0.818 & 0.905 & 0.655 & 0.690 & 0.669 & 0.669 & 0.737 \\
\texttt{Llama-3} (reranking) & 0.741 & 0.805 & 0.904 & 0.647 & 0.666 & 0.640 & 0.662 & 0.723 \\ \cdashline{1-9}[.4pt/1pt]
$\texttt{MonoBERT}_{\texttt{Zero-shot}}$ & 0.581 & 0.660 & 0.703 & 0.552 & 0.629 & 0.575 & 0.623 & 0.618 \\
$\texttt{MonoBERT}_{\texttt{Fine-tuned}}$ & 0.720 & 0.733 & 0.781 & 0.588 & 0.658 & 0.649 & 0.663 & 0.685 \\
\texttt{+BJLM} & 0.691 & 0.758 & 0.749 & 0.617 & 0.648 & 0.646 & 0.646 & 0.680 \\
\texttt{+BALM} & 0.618 & 0.688 & 0.792 & 0.621 & 0.669 & 0.675 & 0.647 & 0.673 \\
\texttt{+MJLM} & 0.593 & 0.754 & 0.801 & 0.606 & 0.648 & 0.659 & 0.659 & 0.674 \\
\texttt{+MALM} & 0.663 & 0.762 & 0.669 & 0.587 & 0.637 & 0.635 & 0.635 & 0.655 \\ \cdashline{1-9}[.4pt/1pt]

$\texttt{ColBERT}_\texttt{Zero-shot}$ & 0.790 & 0.842 & 0.890 & 0.682 & 0.735 & 0.628 & 0.780 & 0.764 \\ 
$\texttt{ColBERT}_\texttt{Fine-tuned}$ & 0.770 & 0.859 & 0.907 & 0.729 & \textbf{0.753} & \textbf{0.670} & \textbf{0.810} & \textbf{0.785} \\
\texttt{+BJLM} & 0.763 & 0.860 & \textbf{0.911} & \textbf{0.744} & 0.679 & 0.545 & 0.778 & 0.754 \\
\texttt{+BALM} & \textbf{0.796} & \textbf{0.878} & 0.912 & 0.737 & 0.707 & 0.597 & 0.790 & 0.774 \\
\texttt{+MJLM} & 0.756 & 0.840 & 0.906 & 0.737 & 0.684 & 0.560 & 0.781 & 0.752 \\
\texttt{+MALM} & 0.743 & 0.851 & 0.905 & 0.742 & 0.661 & 0.656 & 0.774 & 0.762 \\

\bottomrule
\end{tabular}

\caption{Cross-dialect retrieval results on WikiDIR languages and averages over all languages in terms of nDCG@10. Queries are written in German and documents are written in dialects. \textbf{Bold}: Best result for each language.}
\label{tab:main_results}
\end{table*}

\begin{table*}[ht!]
\centering
\small 
\begin{tabular}{l c c c c c c c}
\toprule
 Model & Variations & \textit{als} & \textit{bar} & \textit{nds} & \textit{pfl} & \textit{ksh} & \textbf{avg} \\
 \midrule
\multirow{2}{*}{\texttt{BM25}} & \xmark & 0.401 & 0.483 & 0.371 & 0.293 & 0.401 & 0.390 \\
 & \cmark & 0.357 & 0.453 & 0.347 & 0.277 & 0.325 & 0.353 \\ \cdashline{1-8}[.4pt/1pt]
 
\multirow{2}{*}{\texttt{MonoBERT}} & \xmark & 0.407 & 0.402 & 0.307 & 0.285 & 0.354 & 0.351 \\
 & \cmark & 0.369 & 0.387 & 0.291 & 0.277 & 0.300 & 0.325 \\ \cdashline{1-8}[.4pt/1pt]
 
\multirow{2}{*}{\texttt{ColBERT}} &\xmark & 0.490 & 0.469 & 0.412 & 0.422 & 0.349 & 0.428 \\
 & \cmark & 0.475 & 0.460 & 0.399 & 0.385 & 0.336 & 0.411 \\
\bottomrule
\end{tabular}

\caption{Cross-dialect retrieval results on WikiDIR in terms of nDCG@10. We evaluate BM25 and the fine-tuned versions of MonoBERT and ColBERT on two versions of the WikiDIR analysis split. The first version disregards dialect variations (\xmark) and considers any document relevant that contains the query. The second version considers any document as relevant if it contains the query or any of its dialect variations (\cmark).}
\label{tab:analysis_results}
\end{table*}

\begin{table}
\centering
\adjustbox{max width=\linewidth}{%
\begin{tabular}{lrr}
\toprule
Subdialect & \#Articles & nDCG@10 \\
\midrule
\multicolumn{3}{l}{\textit{Alemannic dialect groups}} \\ 
\midrule
Swabian & 891 & 0.831 \\
Low Alemannic & 3362 & 0.777 \\
Central Alemannic & 221 & 0.538 \\
High Alemannic & 1467 & 0.800 \\
Highest Alemannic & 89 & 0.547 \\ \midrule
 \multicolumn{3}{l}{\textit{North Frisian dialect groups}}  \\ \midrule
Insular North Frisian & 6980 & 0.611 \\ 
Mainland North Frisian & 1825 & 0.485 \\ \midrule
 \multicolumn{3}{l}{\textit{Bavarian dialect groups}}  \\ \midrule
Central Bavarian & 4461 & 0.599 \\
South Bavarian & 297 & 0.612 \\
North Bavarian & 89 & 0.784 \\
\bottomrule
\end{tabular}
}
\caption{CDIR results of BM25 on dialect subgroups.}
\label{tab:subdialect-cdir}
\end{table}

\section{Results}
In the following, we first compare different retrieval paradigms on CDIR (RQ1, RQ2), followed by a detailed analysis on the impact of dialect variance (RQ3). We then discuss the effectiveness of the document translation approach (RQ4). 
We evaluate our results with the normalized discounted cumulative gain of the top-10 ranked items (nDCG@10), as implemented in the ir\_measures framework \citep{DBLP:conf/ecir/MacAvaneyMO22a}.

\subsection{Main Results} 
Table~\ref{tab:main_results} shows our main results. We find that BM25 already performs competitively, achieving an average nDCG@10 score of 0.737. Between both neural IR models, we find that only ColBERT, when fine-tuned on the respective training portions of CDIR ($\texttt{ColBERT}_\texttt{Fine-tuned}$), outperforms BM25. When used in the context of zero-shot cross-lingual transfer ($\texttt{MonoBERT}_\texttt{Zero-shot}$ and $\texttt{ColBERT}_\texttt{Zero-shot}$), both models perform worse. Similarly, our LLM-based baseline, which also has not been fine-tuned on WikiDIR data, fails to outperform BM25. Overall, we find that on average $\texttt{ColBERT}_\texttt{Fine-tuned}$ outperforms other retrieval models (\textbf{RQ1}). We hypothesize that this is due to the fact that its performance is not capped by a (lexical) first-stage retriever. Further specializing retrieval models through bilingual (\texttt{BJLM}, \texttt{BALM}) and multilingual (\texttt{MJLM}, \texttt{MALM}) continual pretraining can bring improvements for the dialects with larger Wikipedias (\textbf{RQ2}).

\subsection{Dialect Variation}
We now investigate dialect variation on the lexical level. The results reported in Table~\ref{tab:analysis_results} are computed on the analysis split. As before, queries are written in German and documents are written in different dialects. Compared to our results on the test split, we find that the results are overall substantially lower. 
This is because titles in dialect articles,\footnote{We used dialect articles to compute monolingual relevance labels (see Section~\ref{sec:pipeline}).} which are known to have different variations, are more likely to be different from their German counterpart. 
All three models consistently perform worse when we include relevant documents that contain dialect spelling variants (\cmark).
The lexical dialect gap is larger for \texttt{BM25} and \texttt{MonoBERT}. This is expected, since lexical methods are limited to exact term matches between queries and documents. Furthermore, our neural models have not been explicitly trained to match different dialect-specific spelling variants. 

We will now investigate CDIR results where the queries have been grouped into their dialect subgroups, as shown in Table~\ref{tab:subdialect-cdir}. Here, too, we find large differences in CDIR performance between different subgroups. This demonstrates the high dialectal variation among subdialects (\textbf{RQ3}).  

\begin{figure}[t!]
    \centering
    \includegraphics[width=\columnwidth]{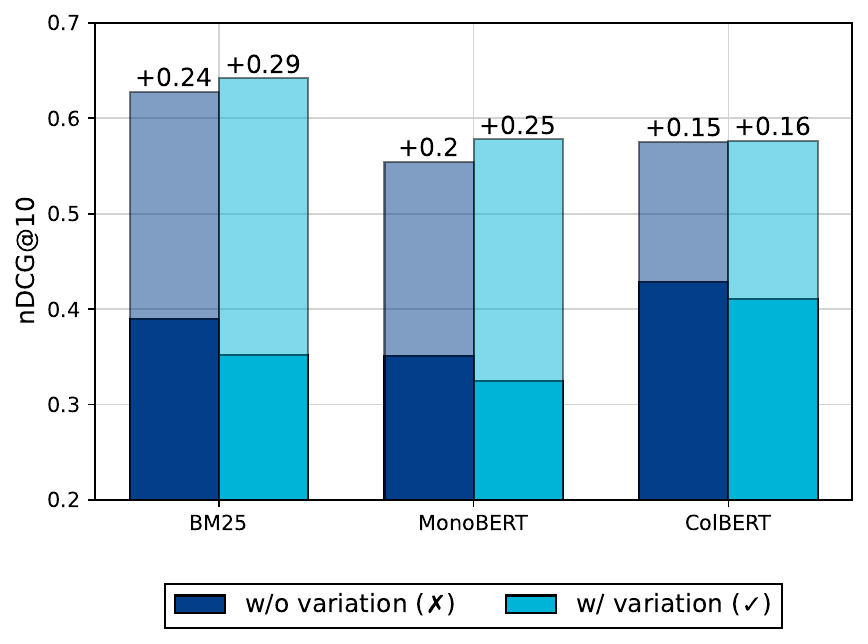}
    \caption{CDIR results on the original training and test data and results (gains) obtained from document translation. We report the average over five dialects.}
    \label{fig:doc-translation}
\end{figure}

\subsection{Document translation}
In this section, we compare the performance between models trained on WikiDIR against models that were trained (and evaluated) on documents, which have been translated into German, as described in Section~\ref{sec:document-translation}. In Figure~\ref{fig:doc-translation}, we compare the average results obtained with documents written in dialects against documents translated to German. We find that document translation yields large gains across all languages (see also Table~\ref{tab:translated-results} in Appendix~\ref{sec:appendix-results}), with gains ranging from +0.15 to +0.29 nDCG@10. Unsurprisingly, BM25 benefits the most among our three models. This shows that document translation into a language with a standard orthography (German) is an effective way of reducing the dialectal orthographic variance (\textbf{RQ4}). 

We now compare the performance with (\cmark) and without (\xmark) dialect variations. On the original data, we find that dialect variations (\cmark) cause a drop in performance. However, on the translated documents we observe the opposite. We hypothesize that this related to the larger number of documents with exact matches (see Table~\ref{tab:fraction-exact-matches} in Appendix~\ref{sec:appendix-results}). That is, in the absence of the lexical dialect gap, treating more documents as relevant (i.e., increasing the recall base) simplifies the retrieval task. 

Our results in Figure~\ref{fig:doc-translation} and Table~\ref{tab:translated-results} suggest that document translation is an effective way to reduce the dialect gap. However, it is important to note that Llama-3 has been exposed to Wikipedia during pretraining. This means that the model has also been exposed to lexical shortcuts (see Section~\ref{sec:pipeline}) and pretrained on dialect code-switched data. Finally, our qualitative analysis reveals two important shortcomings (see Table~\ref{tab:qualitative-analysis}). First, they indicate that employing Llama-3 for dialect translation tends to be more effective in handling orthographic variations linked to different regional pronunciations and sometimes struggles with lexical variations caused by regional word choices. Also, when combined with BM25, the bag-of-words assumption can lead to many false positives.

\section{Conclusion}
In this work, we introduce WikiDIR, a cross-dialect information retrieval (CDIR) test collection spanning seven German dialects. We additionally release dialect dictionaries containing different spelling variants, facilitating a detailed analysis of their impact on CDIR. Our results show that there is a substantial gap between retrieving documents that contain the query and documents containing the query or any of its dialect variations. We also show that, even within a specific dialect, there are large differences in retrieval performance among its subdialects. WikiDIR paves the way for future work on retrieval models that generalize across lexical variants as well as dialect-specific lexical analyzers (e.g. stemmers, lemmatizers). 

\section{Limitations}
In this work, we focused on lexical dialect variation. This neglects the semantic aspect and the information asymmetry between dialects and standard German. Furthermore, our approach for extracting dialect variation candidates (Section~\ref{sec:dialectvariatoin}), which we then annotate to build dialect dictionaries, relies on the connectivity of Wikipedias \citep{arora2024orphan}. In future work, we plan to investigate other approaches of extracting dictionaries without relying on links between articles. 

Wikipedias in low-resource languages can be more prone to quality issues \cite{tatariya2024goodwikipedia}.
Almost all of the wikis included in this study are in the best-quality tier determined by \citeposs{tatariya2024goodwikipedia} set of heuristics (with the exception of \textit{pfl}; tier 3 of~4).
However, this might be an over-estimation due to the quality heuristics favouring wikis with many unique words (which might be inflated due to spelling variation within dialect wikis; cf.\ Section~\ref{sec:dialectvariatoin}).
In terms of the collaborative depth \cite{wikipedia-meta-depth}, the wikis in this study range mostly in the middle, according to both Wikipedia's Depth measure and \citeposs{alshahrani-etal-2023-depth} DEPTH+ measure.

Finally, due to resource constraints, we did not fine-tune Llama-3 for dialect reranking.

\section*{Acknowledgments}
We thank the members of the MaiNLP lab for their insightful feedback on earlier drafts of this paper. We specifically thank Marwin Hättrich for the initial discussion of this work, and Fabian David Schmidt, Philipp Mondorf, Miriam Winkler and Michael A. Hedderich for their help with annotating dialect variations. We thank the anonymous reviewers for their insightful comments. This research is supported by the ERC Consolidator Grant DIALECT 101043235.

\bibliography{anthology,custom}

\appendix

\clearpage
\onecolumn
\section{Results for Translated Documents}
\label{sec:appendix-results}

\begin{table}[ht]
\centering 
\small
\begin{tabular}{l c c c c c c c}
\toprule
 Model & Variations & \textit{als} & \textit{bar} & \textit{nds} & \textit{pfl} & \textit{ksh} & \textbf{avg} \\
 \midrule
\multirow{2}{*}{\texttt{BM25}} & \xmark & 0.598 & 0.672 & 0.654 & 0.602 & 0.613 & 0.628 \\
 & \cmark & 0.616 & 0.677 & 0.651 & 0.622 & 0.643 & 0.642 \\ \cdashline{1-8}[.4pt/1pt]
 
\multirow{2}{*}{\texttt{MonoBERT}} & \xmark & 0.563 & 0.491 & 0.598 & 0.583 & 0.538 & 0.554 \\
 & \cmark & 0.586 & 0.518 & 0.605 & 0.598 & 0.582 & 0.578 \\ \cdashline{1-8}[.4pt/1pt]
 
\multirow{2}{*}{\texttt{ColBERT}} &\xmark & 0.619 & 0.597 & 0.554 & 0.588 & 0.519 & 0.575 \\
 & \cmark & 0.631 & 0.597 & 0.560 & 0.553 & 0.539 & 0.576 \\
\bottomrule
\end{tabular}
\caption{Cross-dialect retrieval results on \textbf{translated documents} of WikiDIR in terms of nDCG@10. We evaluate BM25 and the fine-tuned versions of MonoBERT and ColBERT on two versions of the WikiDIR analysis split. The first version disregards dialect variations (\xmark) and considers any document relevant that contains a the query. The second version considers any document as relevant if it contains the query or any of its dialect variations (\cmark).}
\label{tab:translated-results}
\end{table}

\begin{table}[ht]
\centering 
\small
\begin{tabular}{l c c c c c c c}
\toprule
 Document Language & Variations & \textit{als} & \textit{bar} & \textit{nds} & \textit{pfl} & \textit{ksh} & \textbf{avg} \\
 \midrule
 
\multirow{2}{*}{Dialect (original)} & \xmark & 40.7\% & 34.7\% & 24.7\% & 33.6\% & 42.3\% & 35.2\% \\
 & \cmark & 25.8\% & 29.7\% & 21.3\% & 21.6\% & 28.5\% & 25.4\% \\ \cdashline{1-8}[.4pt/1pt]
 
\multirow{2}{*}{German (translated)} & \xmark & 69.7\% & 85.8\% & 76.9\% & 79.4\% & 73.8\% & 77.1\% \\
 & \cmark & 70.5\% & 80.3\% & 74.0\% & 64.6\% & 69.2\% & 71.7\% \\ 
 
\bottomrule
\end{tabular}
\caption{Percentage of documents containing an exact match of the query (analysis split).}
\label{tab:fraction-exact-matches}
\end{table}

\section{Examples of Errors on Translated Documents}
\label{sec:appendix-error-analysis}

\begin{table}[h]
\footnotesize
\begin{tabular}{p{2cm}p{4.1cm}p{7.1cm}}
\toprule
\textbf{Category} & \textbf{German Query (id)} & \textbf{Error Description} \\ \midrule

\multicolumn{1}{p{2.5cm}}{\multirow{3}{=}[-\baselineskip]{Dialect-Specific Variation}} & \textit{Wort} (9695) & The query for the German term "Wort" (eng.: word) generates many false positives as "Wort" occurs in many documents. In the Ripuarian (ksh) reference document, however, the term \textit{Woot} is not translated correctly. \\
 
 & \textit{Rosine} (98086) & The Bavarian dialect term for "Rosine" (eng.: raisin), \textit{Weinberl}, is not translated at all. \\

 & \textit{Weinbau} (49763) & The Alemannic term for "Weinbau" (eng.: winegrowing), \textit{Räbbau}, is referred to with "Rebbaubegriff". \\ 

\midrule

\multirow{3}{=}[-\baselineskip]{Bag-of-Words} & \raggedright \textit{Schlacht bei Kappel} (3766713) & Retrieved als documents contain many false positives with the subwords "Schlacht bei", but they miss the proper name "Kappel" \\

 & \raggedright\textit{Alter Friedhof Speyer} (4213010) & Retrieved pfl documents contain false positives related to the town name "Speyer", but unrelated to "Alter Friedhof" (eng.: old cemetery). \\
 
 &  \raggedright \textit{Bildende Kunst} (714) & The Bavarian term for "Bildende Kunst" (eng.: fine arts), \textit{Buidnde Kunst}, does not appear in translation. The results for this query contain documents with the term "Kunst" (eng.: art), but are not related to fine arts. \\ 
 
 \bottomrule
\end{tabular}
\caption{Different categories of translation errors in the documents, contributing to weak retrieval performance.}
\label{tab:qualitative-analysis}
\end{table}

\end{document}